\documentclass[sigconf]{acmart}

\usepackage{algorithm}
\usepackage{algorithmic}
\usepackage{multirow}
\usepackage{makecell}
\usepackage{pifont} 
\usepackage{subcaption}
\usepackage{graphicx}
\usepackage{pgfplots}

\AtBeginDocument{%
  }

\copyrightyear{2023}
\acmYear{2023}
\setcopyright{acmlicensed}\acmConference[MM '23]{Proceedings of the 31st ACM International Conference on Multimedia}{October 29-November 3, 2023}{Ottawa, ON, Canada}
\acmBooktitle{Proceedings of the 31st ACM International Conference on Multimedia (MM '23), October 29-November 3, 2023, Ottawa, ON, Canada}
\acmPrice{15.00}
\acmDOI{10.1145/3581783.3612327}
\acmISBN{979-8-4007-0108-5/23/10}

\acmConference[ACM MM]{Make sure to enter the correct
  conference title from your rights confirmation emai}{Oct 28--Nov 03,2023}{Ottawa, Canada}

\begin{document}
\begin{sloppypar}

\title{Enhancing Visually-Rich Document Understanding via Layout Structure Modeling}
%

%
\author{Qiwei Li}
\email{qw-line@whu.edu.cn}
\orcid{0009-0007-2531-6537}
\affiliation{%
  \institution{School of Computer Science, Wuhan University}
  \city{Wuhan}
  \state{Hubei}
  \country{China}
}

\author{Zuchao Li$^*$}
\email{zcli-charlie@whu.edu.cn}
\orcid{0000-0003-0436-8446}
\affiliation{%
  \institution{School of Computer Science, Wuhan University}
  \city{Wuhan}
  \state{Hubei}
  \country{China}
}

\author{Xiantao Cai$^*$}
\email{caixiantao@whu.edu.cn}
\orcid{0000-0002-0764-5085}
\affiliation{%
  \institution{School of Computer Science, Wuhan University}
  \city{Wuhan}
  \state{Hubei}
  \country{China}
  }

\author{Bo Du}
\email{dubo@whu.edu.cn}
\orcid{0000-0001-8104-3448}
\affiliation{%
  \institution{School of Computer Science, Wuhan University}
  \city{Wuhan}
  \state{Hubei}
  \country{China}
  }
\author{Hai Zhao}
\email{zhaohai@cs.sjtu.edu.cn}
\orcid{0000-0001-7290-0487}
\affiliation{%
  \institution{Shanghai Jiao Tong University}
  \city{Shanghai}
  \country{China}
  }

\begin{abstract}
  
In recent years, the use of multi-modal pre-trained Transformers has led to significant advancements in visually-rich document understanding. However, existing models have mainly focused on features such as text and vision while neglecting the importance of layout relationship between text nodes. In this paper, we propose GraphLayoutLM, a novel document understanding model that leverages the modeling of layout structure graph to inject document layout knowledge into the model. GraphLayoutLM utilizes a graph reordering algorithm to adjust the text sequence based on the graph structure. Additionally, our model uses a layout-aware multi-head self-attention layer to learn document layout knowledge. The proposed model enables the understanding of the spatial arrangement of text elements, improving document comprehension. We evaluate our model on various benchmarks, including FUNSD, XFUND and CORD, and achieve state-of-the-art results among these datasets. Our experimental results demonstrate that our proposed method provides a significant improvement over existing approaches and showcases the importance of incorporating layout information into document understanding models. We also conduct an ablation study to investigate the contribution of each component of our model. The results show that both the graph reordering algorithm and the layout-aware multi-head self-attention layer play a crucial role in achieving the best performance.

\end{abstract}



\maketitle

\section{Introduction}

Visually-Rich Document Understanding (VRDU)~\cite{xu2020layoutlm} aims to analyze various types of scanned or digital-born documents that have a rich variety of structures and complex formats. It can assist in a wide range of document-related scenarios, such as report/receipt comprehension~\cite{park2019cord}, document classification~\cite{harley2015evaluation}, and document question answering~\cite{mathew2021docvqa}. Nowadays, VRDU has emerged as a crucial research field and has garnered significant attention from both industry and academia due to its numerous applications.

Unlike conventional natural language understanding (NLU), VRDU not only requires text information but also necessitates the integration of the original document structure and visual information to comprehend non-contiguous text positioned differently throughout the document~\cite{xu2020layoutlm,xu2020layoutlmv2}. Given the diversity of document types and their complex structures, it is challenging to fully understand the meaning of text contained within documents featuring varied content, positions, and contexts using solely textual data. Document comprehension requires the combination of multi-modal information in a meaningful and efficient manner so that models can thoroughly explore the meaning and potential logical relationships within documents.

\begin{figure}[t]
    \centering
    \includegraphics[width=1.0\linewidth]{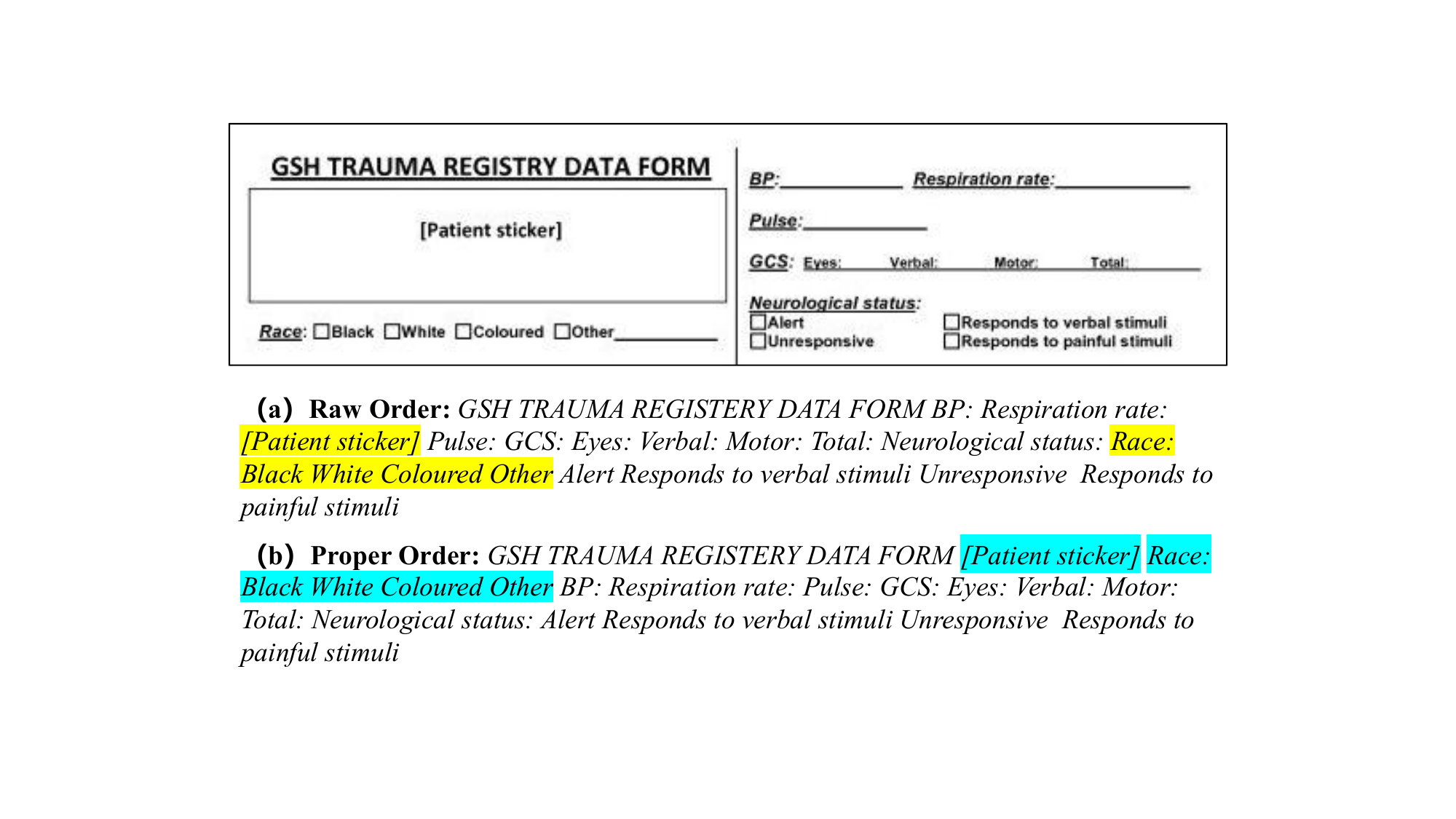}    \caption{Different Reading Order. Raw Order shows fixed text reading order. Proper Order shows a reasonable and organized text reading order.}\label{fig:reading_order}
\end{figure}

In the early stages of VRDU, researchers attempted uni-modal approaches and shallow multi-modal fusion approaches to analyze documents~\cite{yang2016hierarchical,katti2018chargrid,sarkhel2019deterministic,cheng2020one}. These methods relied on pre-trained NLP and CV models separately and combined information from multiple modalities. However, such task-specific approaches required large amounts of annotated data for supervised learning. With recent advancements in pre-training techniques, many more sophisticated document models and pre-training approaches have been proposed. Examples include LayoutLM~\cite{xu2020layoutlm}, LayoutLMv2~\cite{xu2020layoutlmv2}, and BROS~\cite{hong2022bros}, which have made significant progress in the VRDU field. The latest Document AI technique, LayoutLMv3~\cite{huang2022layoutlmv3} and ERNIE-Layout~\cite{peng2022ernie}, are new attempts at multi-modal pre-training and have achieved state-of-the-art results for many VRDU tasks.

Although current research has demonstrated superior performance in the VRDU field, there are still some limitations that need to be addressed. 

\noindent (1) \textbf{Mismatch between the raw order and proper understanding order:} the text scattered in different location boxes of the document are non-continuous. This means that text from different location boxes needs to be joined together before model input. Some works~\cite{xu2020layoutlm,hong2022bros,li2021structurallm,xu2020layoutlmv2,huang2022layoutlmv3} overlook the importance of reading order on the document understanding task and use an unadjusted raw order as the model input. As shown in Figure \ref{fig:reading_order},it is difficult to acquire potential relationship between different text nodes by random or fixed text reading order. This can make it difficult to acquire potential relationships between different text nodes, which can impact machine understanding. A reasonable and organized reading order can reflect the logical or positional relationship between text nodes, making it more appropriate for disambiguation.

\begin{figure}[t]
  \centering
\centering
  \includegraphics[width=1.0\linewidth]{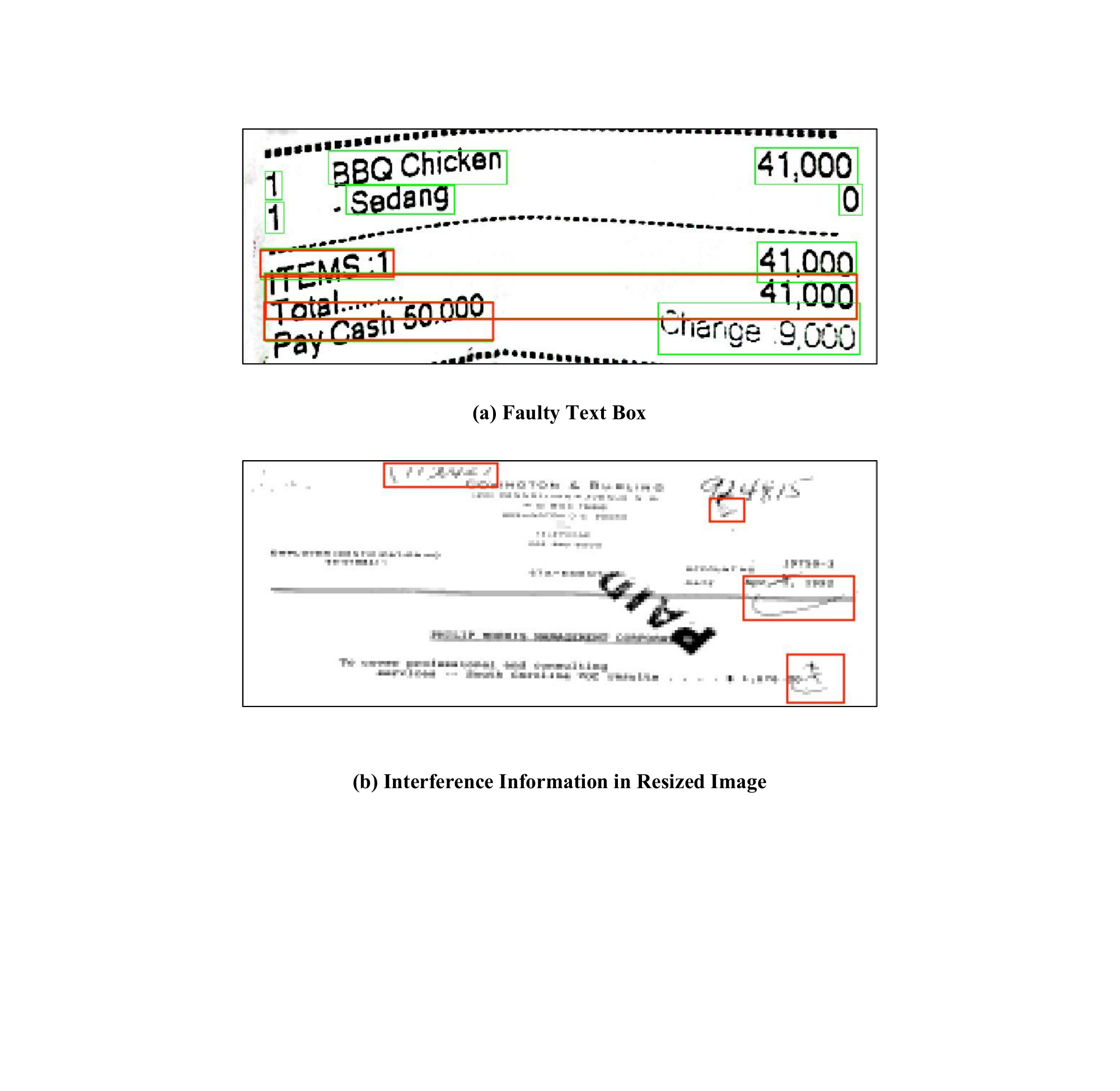} 
  \caption{Fig. (a) shows faulty text box (red box) in low quality document image. Fig. (b) shows interference information (red box) in the resized image.}
  \label{fig:box_and_image}
\end{figure}

\noindent (2) \textbf{Inadequate disclosure of document structure by visual information:} some works~\cite{xu2020layoutlm,hong2022bros,li2021structurallm,xu2020layoutlmv2,gu2022xylayoutlm,huang2022layoutlmv3,peng2022ernie} merely considers multi-modal vision and 2-dimensional (2D) position features to understand document layout . They only utilize multi-modal vision and 2D position features to understand document layout and engage the position and image features in the document understanding process. However, the model needs to learn the relationship between the text nodes only by itself, which can limit the model's understanding of the layout relationship. 
On one hand, low-quality document images can result in faulty text boxes and interference with image information. Thus, relying solely on vision features for document structure reflection is not sufficient. On the other hand, since the input size of the image encoder is limited, the usual practice is to resize the image to a fixed size that meets the requirements. As shown in Figure \ref{fig:box_and_image}, during image preprocessing, LayoutLMv3 resizes the image to $224 \times 224$, which can lead to image blurriness and significant information loss, thus harming the layout understanding.
Moreover, resizing further aggravates the damage caused by faulty text boxes and interference information since it is harder to distinguish them from normal information in a resized image. Additionally, the current use of 2D positions pre-extracted from the original image to indicate document structure is not intuitive. Encoders find it difficult to understand the structure of the document from 2D positions since the document structure is highly understood and related, rather than purely spatial.

In this work, we propose a VRDU pre-trained language model called GraphLayoutLM, which is an improved version of LayoutLMv3 and utilizes a layout relationship graph to improve document understanding. Unlike previous models that focus on providing feature information, GraphLayoutLM concentrates on the position relationship between text nodes and represents this kind of layout relationship using a graph structure. To build GraphLayoutLM, we introduce two optimization strategies: graph sequence optimization and graph mask optimization.

First, we model the document layout information as a hierarchical relational structure graph consisting of multiple text nodes. For example, a paragraph comprises several lines of text, and paragraphs form sections that together make up the entire document. This hierarchical document structure serves as the foundation for constructing the document layout graph. To describe the hierarchical document structure, we pre-build the document layout as a tree. Since siblings usually share a closer relationship than non-siblings, we establish a position relationship between sibling nodes to enrich the tree into a layout graph.

The order of reading documents can vary widely, and a proper sequencing can help us better comprehend the content. To achieve a suitable reading order, we utilize the layout graph built on a tree-based structure that preserves the original hierarchical information. Our approach involves a deep traversal of the layout tree, which arranges closely related nodes adjacent to their relevant position in the sequence. We also sort sibling nodes based on their position information in the graph. As our graph sequence optimization approach, by reordering the graph, we are able to obtain an optimal reading order, which serves as an augmentation strategy to enhance document understanding performance.

Inspired by Graph Attention Networks (GAT)~ \cite{velivckovic2017graph}, we attempt to integrate layout graph information into the VRDU model through a self-attention strategy. GraphLayoutLM introduces a graph-aware self-attention layer that enhances the standard self-attention layers by optimizing the graph mask. This is achieved by incorporating layout graph information through an adjacent matrix. The graph mask selectively masks attention score elements that lack edge relationships between text nodes and inject the relationship between text nodes. This approach enables the model to better focus on the relationship between correlated text nodes, enhancing its performance.

We conduct extensive experiments on typical VRDU tasks to evaluate our proposed GraphLayoutLM model, including three datasets: the FUNSD dataset and the XFUND dataset for form understanding, and the CORD dataset for receipt understanding. The experiments show that GraphLayoutLM achieves a new state-of-the-art (SOTA) performance on these three datasets, significantly outperforming the previous SOTA baseline LayoutLMv3 by 2.29\% and 1.54\% on the FUNSD and XFUND datasets, respectively. These results demonstrate the effectiveness of our proposed GraphLayoutLM. Additionally, we perform ablation studies to evaluate the effectiveness of our optimization strategies. The results show that our proposed graph order optimization and graph mask optimization strategies are both useful for document understanding. Our code will be available at \url{https://github.com/Line-Kite/GraphLayoutLM}.

\section{Related Work}

Self-supervised pre-training technology is rapidly developing and has been successfully applied to the VRDU field. LayoutLM~\cite{xu2020layoutlm} is the first to combine 2D layout information with text and apply the Masked Language Modeling pre-training task to document understanding. Since then, many novel multi-modal pre-training models have been proposed successively, such as LayoutLMv2~\cite{xu2020layoutlmv2}, BROS~\cite{hong2022bros}, StructuralLM~\cite{li2021structurallm} and LayoutLMv3~\cite{huang2022layoutlmv3}. Among them, LayoutLMv3, as a new generation document pre-training model, has achieved excellent performance on multiple datasets.

However, the previously mentioned methods only focus on layout or visual features and fail to consider the reading order of documents. In reality, documents have a specific reading order during reading and writing, which has significant reference value for model learning. LayoutReader~\cite{wang2021layoutreader} is a multi-modal reading order analysis model trained on the large reading sequence dataset ReadingBank, which effectively extracts the correct reading order of a document. However, this complex model often requires a considerable amount of time to predict, which can significantly affect the efficiency of document understanding. XYLayoutLM~\cite{gu2022xylayoutlm} proposes an augmented XY Cut strategy to reorder the input sequence for proper reading order. Besides XYLayoutLM, ERNIE-Layout~\cite{peng2022ernie} also emphasizes the effect of reading order on document understanding and utilizes an advanced document layout analysis toolkit to sort the text input sequence. Unlike the above two model strategies, GraphLayoutLM relies on graph structure to reorder the input sequence. We believe that this method better reflects the logical relationship between text nodes in a simple and training-free way.

Because of the discreteness of the document nodes, it is suitable to represent the layout of the document by graph structure. There have been some previous efforts to apply graph structures to the VRDU field. Liu et al.~\cite{liu2019graph} apply GCN~ \cite{kipf2016semi} to the field of document understanding and make progress in information extraction. Wang et al.~\cite{wang2022ernie} propose ERNIE-mmLayout which is a multi-grained multimodal Transformer to incorporate coarse-grained information into existing pre-trained fine-grained multimodal Transformers based on the graph. ROPE~\cite{lee2021rope} is a new positional encoding technique based on GCN designed to apprehend the sequential presentation of words in documents. FormNet~\cite{lee2022formnet}  construct Super-Tokens for each word by embedding representations from their neighboring tokens through GCN. Limited by space, we will compare them with our method in Appendix \ref{appendix:graph_model_comparison}.

\begin{figure*}[t]
    \centering
    \includegraphics[width=0.9\linewidth]{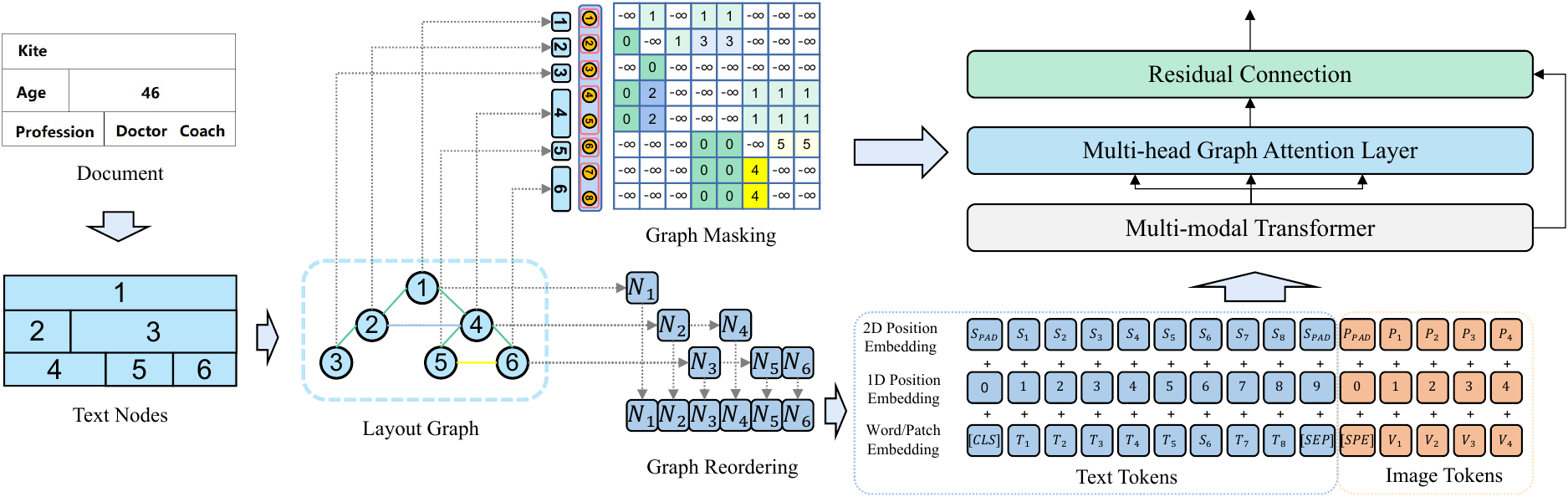}
    \centering
    \caption{The architecture of GraphLayoutLM. In the text extraction stage, the document hierarchy needs to be extracted to generate the document layout graph. The reading order of text tokens sequence is adjusted by Graph Reordering before entering the Multi-modal Transformer.In the output part of Multi-modal Transformer, we use residual connection to connect a multi-head self-attention layer, so as to add the graph mask information for relationship learning.}
    \label{fig:graphlayoutlm}
\end{figure*}

\section{GraphLayoutLM}

We propose a novel GraphLayoutLM, which is based on the currently strongest model, LayoutLMv3. An overview of our proposed GraphLayoutLM can be seen in Figure \ref{fig:graphlayoutlm}. As shown in the figure, text is often scattered throughout different positions in a document. When analyzing a document image using an OCR tool, we can obtain not only the text and its corresponding position information but also the layout level of the entire document, such as sections, paragraphs, and lines. By utilizing the layout level and position information, we can construct a graph that represents the layout of the entire document. GraphLayoutLM builds the graph optimization sequence and graph mask through the relationship graph, which are then used in the input and Transformer encoding. Compared to LayoutLMv3, GraphLayoutLM understands documents through a more logical input sequence and learns the relationship information between text nodes more directly.

\subsection{Basis Architecture}

\subsubsection{Multi-modal Embeddings} 

Consistent with LayoutLMv3, we converted text, position, and image inputs into text embeddings and image embeddings before being fed into the multi-modal Transformer. The text embedding is composed of the text itself, along with 1D and 2D positional information. The 1D position refers to the index of the text in the sequence, while the 2D layout position represents the border coordinates of the text. After the embedding process, these three components are added together to form the text embedding. Meanwhile, the image embedding follows the same approach as ViT~\cite{dosovitskiy2020image} for processing images. In LayoutLMv3, the document image is first adjusted to a size of $H \times W$. If the image is represented by $I \in \mathbb{R}^{C \times H \times W}$, where $C$, $H$, and $W$ represent the channel size, width, and height, respectively, the image is divided into uniform sequences of $P \times P$ blocks. These image blocks are linearly projected into vectors of dimension $D$ and flattened into vector sequences of length $M=HW/P^2$. Finally, the text embedding and image embedding are concatenated and input into the following multi-modal Transformer encoder.

\subsubsection{Multi-modal Transformer} 

In line with the LayoutLMv3 approach, we make some adjustments to the multi-modal Transformer in GraphLayoutLM. Firstly,

we use $\mathbf{b}^{(1D)}$, $\mathbf{b}^{(2D_x)}$, and $\mathbf{b}^{(2D_y)}$ to represent the learnable 1D and 2D position biases which vary across attention heads but shared in all encoder layers.Using $(x_i, y_i)$ as the anchor for the top-left corner of the $i$-th bounding box,we can compute the spatial-aware attention score:
\begin{equation}
a_{i,j}'=a_{i,j}+\mathbf{b}^{(1D)}_{j-i}+\mathbf{b}^{(2D_x)}_{x_j-x_i}+\mathbf{b}^{(2D_y)}_{y_j-y_i},
\end{equation}
where $a_{ij}$ is the  original attention score.
Moreover, we follow CogView~\cite{2021CogView} and modify the attention calculation formula for self-attention:
\begin{equation}
\textrm{Softmax}\left(\frac{\mathbf{Q}^T\mathbf{K}}{\sqrt{d}}\right)  =\textrm{Softmax}\left(\left(\frac{\mathbf{Q}^T}{\alpha\sqrt{d}}\mathbf{K} - \max\left(\frac{\mathbf{Q}^T}{\alpha\sqrt{d}}\mathbf{K}\right)\right)\times \alpha\right),
\end{equation}
where $\alpha$ is scaling factor.

\subsubsection{Pre-training Objectives} 

Our primary goal is to enhance VRDU performance through layout graphs, we still adopt three types of unsupervised pre-training methods: MLM, MIM, and WPA, similar to those used in LayoutLMv3. By incorporating these techniques, our complete pre-training objective is defined as $L = L_{MLM} + L_{MIM} + L_{WPA}$.The detailed introduction of pre-training objectives is shown in Appendix \ref{appendix:layoutlmv3_pretrain}.

\subsection{Layout Graph Modeling}\label{sec:graph_modeling}

In GraphLayoutLM, we treat each basic text node as a segment-level entity, representing a line of text. The entire document is comprised of these text lines, which can be described as a set of nodes $N=\{n_1,...,n_m\}$. However, not all text nodes in these lines are correlated. To find associations between texts and avoid redundant connections, we extract document hierarchy information such as chapters and paragraphs using OCR tools or layout information from the data. For instance, consider a paragraph that contains text lines which can be defined as $P=\{n_1,...,n_k\}$, where $P$ is the paragraph region. We determine the text node representing the whole paragraph, $n_p$, based on its 2D position:
\begin{equation}
L(n_i) = \left\{
	\begin{aligned}
        & 1 ,\quad box_i=Top(P) \quad and \quad box_i=Left(P)\\
	& 0 ,\quad otherwise\\
	\end{aligned}
	\right
	.,
\end{equation}

where the function $L(n_i)$ determines whether or not $n_i$ can represent the entire paragraph, i.e., $n_p = n_i$. This determination is based on 2D position information of $n_i$, denoted as $box_i$, as well as the 2D position information of the top-most node and left-most node in $P$, represented by $Top(P)$ and $Left(P)$, respectively. This function is designed to mimic the way in which people typically write documents.

After determining the representation text node $n_p$ of the paragraph $P$, we connect $n_p$ with other nodes in the paragraph except for $n_p$ with "\textrm{\textit{Parent-Child}}" relationship to construct a subtree $T_p$, which is formulated as follows:
\begin{equation}
T_p = <n_p,\complement_{P}{n_p}, \textrm{\textit{Parent-Child}}>,
\end{equation}
where $\complement_{P}{n_p}$ means the complement set of $n_p$ in $P$.

Using the method described above, we can obtain a set of subtrees from the document, denoted as $T=\{T_1,...,T_n\}$. However, to obtain complete trees instead of forests, we need to introduce a pseudo-global root node, $n_g$, which connects all the root nodes of the subtrees. By doing this, we can create a document layout tree, $T_g$.

The layout tree reflects the hierarchical structure of the document, with parent nodes and child nodes. However, the relative spatial relationships between nodes within a level are also an important aspect of the document's structural information and are useful for understanding the document.
To fully utilize the relative spatial information of text nodes, we compare the positions of sibling nodes in the document tree $T_g$ to determine their spatial relationship.

\begin{figure}[t]
    \centering
    \includegraphics[width=0.9\linewidth]{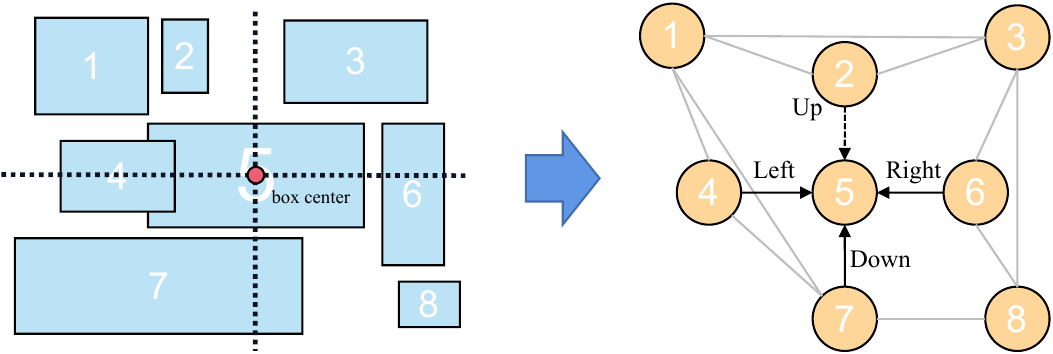}
    \centering
    \caption{An example to establish the position relationship of sibling nodes.}
    \label{fig:sibling_nodes_relationship}
\end{figure}

Figure \ref{fig:sibling_nodes_relationship} shows the process of establishing the position relationship of sibling nodes. Suppose that the 2D position of a node $n_i$ is $n_i.box=[x^i_1,y^i_1,x^i_2,y^i_2]$, where $(x^i_1,y^i_1)$ and $(x^i_2,y^i_2)$ represents the coordinates of the two points in the lower left and upper right corners of the rectangular region in which node $n_i$ is located. We take the center point of $n_i.box$ of node $n_i$ as the origin, and draw horizontal judgment line $l^i_x=(x^i_1+x^i_2)/2$ and vertical judgment line $l^i_y=(y^i_1+y^i_2)/2$. In the comparison process, if the 2D position $n_j.box$ of node $n_j$ intersects $l^j_x$ or $l^j_y$, the relationship will be established according to its 2D position, which can be expressed as:
\begin{equation}
\resizebox{0.9\hsize}{!}{$\begin{aligned}
rel(n_i,n_j)=\left\{
	\begin{aligned}
        & Left,\quad l^i_x \ intersects \ n_j.box \ on \ the \ left \ side\\
        & Right,\quad l^i_x \ intersects \ n_j.box \ on \ the \ right \ side\\
        & Up,\quad l^i_y \ intersects \ n_j.box \ at \ the \ top\\
        & Down,\quad l^i_y \ intersects \ n_j.box \ at \ the \ bottom\\
        & None ,\quad otherwise\\
	\end{aligned}
	\right
	..
 \end{aligned}$}
\end{equation}
Note that if the relationship between two nodes is unidirectional, we can make it bidirectional by adding the inverse relationship that is based on the original unidirectional relationship.

Complded by positional relationships, the document relationship tree $T_g$ will be enriched into the document graph structure $G$:
\begin{equation}
G=T_g\cup\{<n_i,n_j,rel(n_i,n_j)>|n_i,n_j \in T_g.sibling\},
\end{equation}
where $T_g.sibling$ means the set of sibling nodes in $T_g$.

\subsection{Graph Reordering Strategy}

By modeling the layout as a graph, we can optimize the input sequence of text to facilitate document understanding. However, some existing approaches, such as XYLayoutLM~\cite{gu2022xylayoutlm}, focus solely on the relative position in the 2D position layout, ignoring the hierarchy that can be captured from visual information.

Our layout graph is designed to capture comprehensive spatial relationships, which allows for greater flexibility in optimizing reading sequences. This is especially important because different people read documents in different orders, so a reasonable reading order may vary. Additionally, we believe that the reading order extracted from our document layout graph can help the model identify potential associations between text nodes, improving overall comprehension. We show a reordering example in Appendix \ref{appendix:reordering}. The overall graph reordering strategy is shown in Algorithm \ref{alg:graph_reorder}.

\noindent\textbf{Hierarchical Reordering}: As discussed in subsection \ref{sec:graph_modeling}, the document relationship graph consists of a layout document tree $T_g$, which represents the hierarchical structure of the document layout. The hierarchical reordering involves a depth-first traversal of the entire relational structure tree starting from the root node. It is important to note that the root node holds no meaningful information and should be removed from the sequence during input sequence construction. During the traversal, nodes are arranged in order of their parent, child, and finally sibling nodes. This depth-first traversal ordering strengthens connections between nodes and their adjacent regions, as sibling nodes are always located within the same region.

\noindent\textbf{Sibling Reordering}: Hierarchical reordering optimizes the input sequence of text based on its hierarchical relationship. However, to properly order sibling nodes that share the same parent, we need an additional rule to adjust their sequence order at the same level. For this purpose, we use box position information of text nodes to rank the text sequence at the same level according to the position relationship between text nodes. The logic behind position reordering is to sort sibling nodes from top-to-bottom and left-to-right. Specifically, text nodes located at the top of the document are preferred. If the vertical position is roughly the same, then horizontal position is compared, with the left side taking preference. This reordering method is more in line with human habits of writing and reading documents, enabling the model to better understand the logical relationship between text nodes at the same level.

\begin{algorithm} 
	\caption{Graph Reorder Algorithm} 
    \small
	\label{alg:graph_reorder} 
	\begin{algorithmic}
            \REQUIRE $G$ 
		\ENSURE $order=Reorder(G)$
            \STATE $order \gets empty$
            \STATE $stack \gets empty$
            \STATE $stack.push(G.root)$
            \WHILE{$stack$ is not empty} 
            \STATE $node \gets stack.pop()$
            \STATE $order.put(node)$
            \STATE $sort \gets empty$
            \WHILE{$node$.$children$ is not empty}
            \STATE $child \gets node.children$
            \STATE $index \gets 0$
            \WHILE{index is not out of the range of $sort$}
            \STATE $temp \gets sort[index]$
            \IF{$temp$ at the top or left of $child$}
            \STATE $index \gets index+1$
            \ELSE 
            \STATE break
            \ENDIF
            \ENDWHILE
            \STATE $sort.insert(index,child)$
            \ENDWHILE
            \WHILE{$sort$ is not empty}
            \STATE $child \gets sort.pop()$
            \STATE $stack.push(child)$
            \ENDWHILE
            \ENDWHILE
	\end{algorithmic} 
\end{algorithm}

\subsection{Graph Masking Strategy}

The document layout graph contains rich and clear information about the document structure. In addition to optimizing the order of text input, it can improve document representation learning. The graph presents the position and hierarchical relationships among its text nodes in the form of edges, which can reveal logical connections between nodes. For instance, in a typical Q\&A format, the answer text node typically appears on the right or lower side of the question text node. To incorporate this graph information into the document representation learning, we propose a novel graph masking strategy.

Figure \ref{fig:graphlayoutlm} illustrates the bijection but surjective mapping between text tokens ($T=\{t_1,...,t_l\}$) and text nodes ($N=\{n_1,...,n_m\}$). To implement graph masking in Transformer encoding, we convert the relationships between nodes in a graph to relationships between text tokens. Specifically, any text token $t_i \in n_a$ in node $n_a$ and any text token $t_j \in n_b$ in node $n_b$ have the same relationship $rel(t_i, t_j) = rel(n_a, n_b)$ as their corresponding nodes. This completes the mapping from relationships between nodes to relationships between text tokens.

Although the document is recognized by OCR and linearized into a language sequence, its sequence differs from that of natural language. Tokens placed in close proximity may not necessarily have strong correlations; they are likely arranged that way due to linearization. Conversely, tokens that are far apart may not have smaller relationships; they may only appear distant because there are many text nodes at the same level. Therefore, to directly reflect the document's structure from an encoding perspective, we use a graph masking strategy. Specifically, we propose an improved Graph-aware Transformer layer that calculates the self-attention score of the text representation and adds a graph mask $M_g$, containing relational information, in the calculation process. This drives the model to pay more attention to relevant parts of the layout and ignore irrelevant ones, thereby reducing the impact of irrelevant information on understanding. The model becomes aware of document structures as a result. For the graph mask $M \in \mathbb{R}^{n \times n}$, element $m_{i,j}$ of $M$ is computed by:
\begin{equation}
m_{i,j} = \left\{
	\begin{aligned}
        & w_{edge}(t_i,t_j) ,\quad  edge(t_i,t_j) \in G\\
	& -9e15 ,\quad edge(t_i,t_j) \notin G\\
	\end{aligned}
	\right
	.,
\end{equation}
where $w_{edge}(t_i,t_j)$ is the weight of relation labels between tokens $t_i$ and $t_j$. 

In our graph-aware Transformer, it also employs multi-head self-attention for text encoding, in which the self-attention score $e_{i,j}$ between two tokens $t_i$ and $t_j$  can be formulated as:
\begin{equation}
e_{i,j} = \frac{\textbf{W}_Q h_i (\textbf{W}_K h_j)^T}{\sqrt{d_k}}+m_{i,j},
\end{equation}
where $\textbf{W}_Q$ and $\textbf{W}_K$ are learnable projection weights for self-attention, $h_i$ and $h_j$ are token representation input for tokens $t_i$ and $t_j$ respectively.

Because of the existence of graph masking, the model will selectively ignore the elements without edge relationship in the process of self-attention learning, and then inject the layout structure into text representations. The final text representations $H_g$ calculated by graph-aware self-attention with layout is as follows:
\begin{equation}
H_g = \text{Linear}(\text{Softmax}(\frac{\textbf{W}_Q H (\textbf{W}_K H)^T}{\sqrt{d_k}} + M_g) \cdot \textbf{W}_V H),
\end{equation}
where $H$ is the input text representations, $W_V$ is also the learnable value projection weight.

In order to prevent potential information loss caused by masking resulting from layout graph construction errors, we use residual connections to merge the input text representation with the output representation that is aware of the layout graph. Residual connections are useful because they allow us to retain information from the input representation and combine it with the output representation in a way that improves model performance. Ultimately, this combination results in a final output that can be used for downstream tasks:
\begin{equation}
H_{output}= \textsc{Residual}(H, H_g).
\end{equation}

\section{Experiments}

\subsection{Experimental Setup}

\noindent\textbf{Dataset.} We conduct form and receipt understanding experiments on several widely-used VRDU datasets, including FUNSD~\cite{jaume2019funsd} and CORD~\cite{park2019cord} following LayoutLMv3. FUNSD is a form understanding dataset for scanned documents that contains 199 annotated documents (149 in the training set and 50 in the test set) with a total of 9,707 semantic entity labels. Our focus is on the semantic entity labeling task, where each entity is assigned one of four labels: "question," "answer," "header," or "other." Similarly, CORD is a receipt understanding dataset consisting of 1,000 receipts split into 800 training, 100 validation, and 100 test examples. The dataset defines 30 semantic labels under four broad categories for the receipt key information extraction task.

Furthermore, we conduct additional experiments on the XFUND~\cite{xu2022xfund} dataset, which is a manually annotated multilingual document understanding dataset containing 199 documents in seven languages. While XFUND also includes annotations in multiple languages, we focus specifically on the Chinese versions due to pre-training limitations. For the Chinese version of XFUND, it includes 13,857 semantic entity annotations. Like FUNSD, this dataset is divided into a 149-sample training set and a 50-sample test set, and our task is to label each semantic entity as "question," "answer," "header," or "other." 

\begin{table*}[t]
    \centering
    \caption{Results on FUNSD and CORD test sets. “T/L/I/G” denotes “text/layout/image/graph” modality. Results for LayoutLMv3 that are labeled with "$^\dagger$" indicate that the results are obtained through replication in our experiments.}
    \label{tab:sota}
    \begin{tabular}{lllcccccc}
    \toprule
    \multirow{2}{*}{\bf Model} & \multirow{2}{*}{\bf Parameters} & \multirow{2}{*}{\bf Modality} &\multicolumn{3}{c}{\bf FUNSD} &\multicolumn{3}{c}{\bf CORD}   \\
     & & & \bf Precision & \bf Recall  & \bf F1 & \bf Precision & \bf Recall & \bf F1 \\
     \midrule
     $\textrm{BERT}_{\rm BASE}$~\cite{devlin2018bert} & 110M & T & 54.69 & 67.10 & 60.26  & 88.33 & 91.07 & 89.68\\
     $\textrm{RoBERTa}_{\rm BASE}$~\cite{liu2019roberta} & 125M & T  & 63.49 & 69.75 & 66.48 & - & - &  93.54\\
     $\textrm{BROS}_{\rm BASE}$~\cite{hong2022bros} & 110M & T+L & 81.16 & 85.02 & 83.05  & 95.58 & 95.14 & 95.36\\
     $\textrm{LayoutLM}_{\rm BASE}$~\cite{xu2020layoutlm} & 160M & T+L+I & 76.77 & 81.95 & 79.27  & 94.37 & 95.08 & 94.72\\
     $\textrm{XYLayoutLM}_{\rm BASE}$~\cite{gu2022xylayoutlm} & - & T+L+I & - & - & 83.35 & - & - & -\\
     $\textrm{LayoutLMv2}_{\rm BASE}$~\cite{xu2020layoutlmv2} & 200M & T+L+I & 80.29 & 85.37 & 82.76 & 94.53 & 95.39 & 94.95\\
     $\textrm{DocFormer}_{\rm BASE}$~\cite{appalaraju2021docformer} & 183M & T+L+I & 80.76 & 86.09 & 83.34 & 96.52 & 96.14 & 96.33\\
     $\textrm{ERNIE-Layout}_{\rm BASE}$~\cite{peng2022ernie} & - & T+L+I & - & - & 90.28 & - & - & 96.61\\
     $\textrm{LayoutLMv3}_{\rm BASE}$~\cite{huang2022layoutlmv3} & 133M & T+L+I & - & - & 90.29 & - & - & 96.56\\
     $\textrm{LayoutLMv3}_{\rm BASE}$~\cite{huang2022layoutlmv3}$^\dagger$ & 133M & T+L+I & 90.82 & 91.55 & 91.19 & 96.35 & 96.71 & 96.53\\
     \bf $\textrm{GraphLayoutLM}_{\rm BASE}$ (Ours) & 135M & T+L+I+G & \textbf{92.46} & \textbf{93.85} & \textbf{93.15} & \textbf{97.02} & \textbf{97.53} & \textbf{97.28}\\
     \midrule
     $\textrm{BERT}_{\rm LARGE}$~\cite{devlin2018bert} & 340M & T& 61.13 & 70.85 & 65.63 & 88.86 & 91.68 & 90.25\\
     $\textrm{RoBERTa}_{\rm LARGE}$~\cite{liu2019roberta} & 355M & T& 67.80 & 73.91 & 70.72 & - & - & 93.80\\
     $\textrm{LayoutLM}_{\rm LARGE}$~\cite{xu2020layoutlm} & 343M & T+L& 75.96 & 82.19 & 78.95 & 94.32 & 95.54 & 94.93 \\
     $\textrm{BROS}_{\rm LARGE}$~\cite{hong2022bros} & 340M & T+L& 82.81 & 86.31 & 84.52  & - & - & 97.28\\
     $\textrm{StructuralLM}_{\rm LARGE}$~\cite{li2021structurallm} & 355M & T+L& 83.52 & 86.81 & 85.14 & - & - & -\\
     $\textrm{LayoutLMv2}_{\rm LARGE}$~\cite{xu2020layoutlmv2} & 426M & T+L+I & 83.24 & 85.19 & 84.20 & 95.65 & 96.37 & 96.01\\
     $\textrm{DocFormer}_{\rm LARGE}$~\cite{appalaraju2021docformer} & 536M & T+L+I & 82.29 & 86.94 & 84.55 & 97.25 & 96.74 & 96.99\\
     $\textrm{ERNIE-Layout}_{\rm LARGE}$~\cite{peng2022ernie} & - & T+L+I & - & - & 93.12 & - & - & 97.21\\
     $\textrm{LayoutLMv3}_{\rm LARGE}$~\cite{huang2022layoutlmv3} & 368M & T+L+I & - & - & 92.08 & - & - & 97.46\\
     $\textrm{LayoutLMv3}_{\rm LARGE}$~\cite{huang2022layoutlmv3}$^\dagger$ & 368M & T+L+I & 91.51 & 92.70 & 92.10 & 97.45 & 97.52 & 97.49\\
     \bf $\textrm{GraphLayoutLM}_{\rm LARGE}$ (Ours) & 372M & T+L+I+G & \textbf{94.49} & \textbf{94.30} & \textbf{94.39} & \textbf{97.75} & \textbf{97.75} & \textbf{97.75}\\
     \bottomrule
    \end{tabular}
\end{table*}

\noindent\textbf{Model Details.} To ensure a fair comparison, GraphLayoutLM's network architecture follows that of LayoutLMv3. The $\mathrm{GraphLayoutLM_{BASE}}$ model employs a 12-layer Transformer encoder with 12-head self-attention, a hidden size of $D = 768$, and 3,072 intermediate dimensions for feed-forward networks. The number of heads in the graph-aware multi-head self-attention layer is 6. In contrast, the $\mathrm{GraphLayoutLM_{LARGE}}$ model uses a 24-layer Transformer encoder with 16-head self-attention, a hidden size of $D = 1,024$, and 4,096 intermediate dimensions for feed-forward networks. The number of heads in the graph-aware multi-head self-attention layer is 8.
For both models, we set the maximum text sequence length to $L=512$ and the image input size to $C\times H\times W = 3\times 224 \times 224$. The image patch size $P$ is 16, and the visual sequence length $M$ is 196. We also set the Transformer self-attention layer scaling factor $\alpha$ to 32.

To ensure a fair comparison with LayoutLMv3, we adopt similar settings for our model's pre-training, including the data used and the pre-training steps. For fine-tuning on downstream tasks, we set hyper-parameters consistent with those used in LayoutLMv3. The detailed hyper-parameters are shown in Appendix \ref{appendix:parameters}. These measures help ensure that any differences in performance between the models can be attributed to architectural changes rather than variations in training procedures.

\noindent\textbf{Baselines.} We compare our model against several baselines, including typical self-supervised pre-training language models such as BERT~\cite{devlin2018bert} and RoBERTa~\cite{liu2019roberta}, as well as document understanding models that focus on text and layout modalities like BROS~\cite{hong2022bros}, LayoutLM~\cite{xu2020layoutlm}, and StructuralLM~\cite{li2021structurallm}. We also includes multilingual pre-trained models such as XLM-RoBERTa~\cite{conneau2019unsupervised}, LayoutXLM~\cite{xu2021layoutxlm}, and XYLayoutLM~\cite{gu2022xylayoutlm}, as well as popular models that incorporate text, layout, and image modalities like LayoutLMv2~\cite{xu2020layoutlmv2}, DocFormer~\cite{appalaraju2021docformer}, ERNIE-Layout~\cite{peng2022ernie}, and LayoutLMv3~\cite{huang2022layoutlmv3}.

\subsection{Main Analysis}

The results for FUNSD and CORD are presented in Table \ref{tab:sota}. For both base and large model sizes, GraphLayoutLM has demonstrated improved precision, recall, and F1 scores when compared with the baseline model, LayoutLMv3. On the FUNSD dataset, it can be observed that GraphLayoutLM achieved an F1 score of 93.15 with the base size, which is 1.96 higher than the original $\textrm{LayoutLMv3}_{\rm BASE}$ F1 score of 91.19. Moreover, the base size GraphLayoutLM F1 score is also 1.05 higher than the $\textrm{LayoutLMv3}_{\rm LARGE}$ F1 score of 92.10, and even surpasses the previous SOTA, $\textrm{ERNIE-Layout}_{\rm LARGE}$ F1 score of 93.12. The large GraphLayoutLM F1 score (94.39) is higher than the previous SOTA (93.12). These results serve as evidence of our GraphLayoutLM's exceptional performance and its significant improvement over LayoutLMv3.

On the CORD dataset, the $\textrm{GraphLayoutLM}_{\rm BASE}$ F1 score is 97.28, which is 0.75 higher than the $\textrm{LayoutLMv3}_{\rm BASE}$ F1 score of 96.53. The $\textrm{GraphLayoutLM}_{\rm LARGE}$ F1 score is 97.75, which is 0.26 higher than the $\textrm{LayoutLMv3}_{\rm LARGE}$ F1 score of 97.49. It can be observed that GraphLayoutLM has less effect on promotion for CORD. This may be because CORD is a receipt dataset, where document text content is minimal, and the format is simple. Layout graphs are more suitable for aiding models understand complex documents. However, even in this scenario, GraphLayoutLM still provides extra enhancement and has become the new SOTA for the CORD dataset.

In this table, we have also included the number of parameters for each model. It is evident that the GraphLayoutLM base and large size models only have 2M and 4M more parameters than their corresponding LayoutLMv3 models, respectively. This further emphasizes that our strategies can yield significant performance enhancements with just a few additional parameters.

To assess the language-independence of our GraphLayoutLM model, we present the results on the Chinese dataset, XFUND, in Table \ref{tab:xfund}. As only F1 scores are publicly available for the baseline models on XFUND, we compare only the F1 scores in this experiment. Based on comparative experiments, we observe that GraphLayoutLM achieves SOTA with an F1 score of 93.56, which is higher than the original SOTA LayoutLMv3 F1 score of 92.02. These experimental findings demonstrate that GraphLayoutLM showcases good adaptability and delivers excellent performance across multiple languages.

\begin{table}[t]
    \centering
    \small
    \caption{Results on XFUND. The LayoutLMv3 labeled with "$^\dagger$" means that the results are the results of our experiment reproduced.The models labeled with "$^\ddagger$" mean that we get the results from github.}
    \label{tab:xfund}
    \begin{tabular}{llc}
    \toprule
    \multirow{2}{*}{\bf Model} & \multirow{2}{*}{\bf Modality} & \bf XFUND  \\
     & & \bf F1  \\
    \midrule
    $\textrm{XLM-RoBERTa}_{\rm BASE}$~\cite{conneau2019unsupervised} & T & 87.74 \\
    $\textrm{XLM-RoBERTa}_{\rm LARGE}$~\cite{conneau2019unsupervised} & T & 89.25 \\
    $\textrm{LayoutXLM}_{\rm BASE}$~\cite{xu2021layoutxlm} & T+L+I & 89.24 \\
    $\textrm{LayoutXLM}_{\rm LARGE}$~\cite{xu2021layoutxlm} & T+L+I & 91.61 \\
    $\textrm{XYLayoutLM}_{\rm BASE}$~\cite{gu2022xylayoutlm} & T+L+I & 91.76 \\
    $\textrm{ERNIE-LayoutX}_{\rm BASE}$~\cite{peng2022ernie}$^\ddagger$ & T+L+I & 88.58\\
    $\textrm{LayoutLMv3-Chinese}_{\rm BASE}$~\cite{huang2022layoutlmv3}$^\ddagger$ & T+L+I & 92.02\\
    $\textrm{LayoutLMv3-Chinese}_{\rm BASE}$~\cite{huang2022layoutlmv3}$^\dagger$ & T+L+I & 91.82 \\
    \bf $\textrm{GraphLayoutLM-Chinese}_{\rm BASE}$(Ours) & T+L+I+G & \textbf{93.56} \\
    \bottomrule
    \end{tabular}
\end{table}

\subsection{Ablation Study}

\begin{table}[t]
    \setlength{\tabcolsep}{3pt}
    \centering
    \small
    \caption{Ablation Study}
    \label{tab:ablation_study}
    \begin{tabular}{c|cc|cccc}
    \toprule
    \bf Dataset& \bf \makecell{Graph \\ Reorder} & \bf \makecell{Graph \\ Mask} & \bf Accuracy & \bf Precision & \bf Recall & \bf F1 \\
    \midrule
    \multirow{4}{*}{FUNSD} & \ding{55}   & \ding{55} & 84.76  & 90.82 & 91.55 & 91.19  \\
    &\ding{51}   & \ding{55} & 85.70  & 92.36 & 93.15 & 92.75  \\
    &\ding{55}   & \ding{51} & 86.75   & 91.73 & 92.05 & 91.89  \\
    &\ding{51}   & \ding{51} & 88.39  & 92.46 & 93.85 & 93.15  \\
    \midrule
    \multirow{4}{*}{CORD} & \ding{55}   & \ding{55} & 97.11  & 96.35 & 96.71 & 96.53  \\
    &\ding{51}   & \ding{55} & 97.33  & 96.79 & 97.16 & 96.97  \\
    &\ding{55}   & \ding{51} & 97.88   & 96.94 & 97.23 & 97.09  \\
    &\ding{51}   & \ding{51} & 98.01  & 97.02 & 97.53 & 97.28  \\
    \midrule
    \multirow{4}{*}{XFUND} & \ding{55}   & \ding{55} & 85.87  & 89.79 & 93.94 & 91.82  \\
    &\ding{51}   & \ding{55} & 85.61  & 89.98 & 94.37 & 92.12  \\
    &\ding{55}   & \ding{51} & 90.88   & 91.58 & 94.43 & 92.99  \\
    &\ding{51}   & \ding{51} & 91.19  & 91.80 & 95.38 & 93.56  \\
    \bottomrule
    \end{tabular}
\end{table}

We conduct ablation experiments on FUNSD, CORD, and XFUND to test the effectiveness of our proposed graph reordering and graph masking strategies in GraphLayoutLM. We take LayoutLMv3 with a base size as the baseline model, add graph reordering and graph mask methods, respectively, and finally demonstrate the results of GraphLayoutLM combined with the two methods.

Table \ref{tab:ablation_study} shows the ablation experiment results on FUNSD, CORD, and XFUND. It can be seen that the graph reordering method and graph mask method can improve precision, recall, and F1. GraphLayoutLM, which combines the two strategies, performs better. It is worth noting that the graph mask strategy shows a more significant improvement in accuracy (1.99 on FUNSD, 0.77 on CORD, and 5.01 on XFUND). The convergence curves examples are shown in Appendix \ref{appendix:convergence} to verify the credibility of the ablation experiment.

\subsection{Further Analysis}

\begin{table}[t]
    \centering
    \small
    \caption{Different Order.}
    \label{tab:different_order_on_funsd}
    \begin{tabular}{c|cccc}
    \toprule
    \bf Order & \bf Accuracy & \bf Precision & \bf Recall & \bf F1 \\
    \midrule
    Default   & 86.75  & 91.73 & 92.05 & 91.89  \\
    BFS   & 86.95   & 92.15 & 92.11 & 92.13  \\
    DFS   & \textbf{88.39}  & \textbf{92.46} & \textbf{93.85} & \textbf{93.15}  \\
    \bottomrule
    \end{tabular}
\end{table}

\noindent\textbf{Graph Reordering Strategy.} To further study the influence of different graph reordering algorithms, we conduct experiments on the FUNSD dataset using various algorithms. As shown in Table \ref{tab:different_order_on_funsd}, "Default" represents the original order, "BFS" represents the order of breadth-first traversal generation, and "DFS" represents the order of depth-first traversal generation. The results indicate that both "BFS" and "DFS" outperform "Default" in all evaluation indices. Specifically, "BFS" has an accuracy score and f1 score that are 0.20 and 0.24 higher than "Default", respectively, while "DFS" has an accuracy score and f1 score that are 1.64 and 1.26 higher than "Default", respectively. These findings suggest that a certain logical ordering is better than the original ordering. Additionally, "DFS" produces the best results among the algorithms tested, which verifies its superiority on our layout graph because the resulting order is more in line with human habits.

\begin{table}[t]
    \centering
    \small
    \caption{Graph Masking Type.}
    \label{tab:graph_mask}
    \begin{tabular}{l|cccc}
    \toprule
    \bf Masking & \bf Accuracy & \bf Precision & \bf Recall & \bf F1 \\
    \midrule
    LayoutLMv3   & 84.76  & 90.82 & 91.55 & 91.19  \\
    \midrule
    GraphLayoutLM   & \textbf{86.75}  & \textbf{91.73} & \textbf{92.05} & \textbf{91.89}  \\
    w/o Graph Masking   & 82.96  & 90.32 & 91.54 & 90.88  \\
    w/ Hier Masking   & 86.09  & 90.40 & 91.80 & 91.09  \\
    w/ Sib Masking   & 85.60  & 91.77 & 90.85 & 91.31  \\
    \bottomrule
    \end{tabular}
\end{table}

\noindent\textbf{Graph Masking Strategy.} We study the validity of different graph masking strategies, and conduct additional experiments on the FUNSD dataset. The results are presented in Table \ref{tab:graph_mask}. "w/o Graph Masking" indicates that only the self-attention layer is added to the model without any image mask. "w/ Hier Masking" and "w/ Sib Masking" indicate that only the hierarchical graph mask and sibling graph mask are added to the model, respectively. Our analysis result shows that "w/o Graph Masking" doesn't do better than LayoutLMv3 even though it has more parameters. This suggests that the key to performance improvement from our Graph Masking strategy is not due to the increase in the number of parameters. Adding graphs with incomplete relationship information, as done with "w/ Sib Masking," results in little gain for the model and may even have a negative impact on model optimization, while "w/ Hier Masking" lead to similar results. Our proposed GraphLayoutLM structure outperforms all other models in the table, indicating that the improvement in model performance is mainly due to our novel layout graph and graph utilization strategies.

\section{Conclusion}

In this paper, we propose GraphLayoutLM - a layout graph-based pre-trained multi-modal model for document understanding. We focus on the layout structure of documents and have designed a novel layout graph construction method. Based on the layout graph, we propose graph reordering and graph masking strategies. Our experiments show that GraphLayoutLM exhibits superior performance and achieves new state-of-the-art results in the FUNSD, CORD, and XFUND datasets, shedding light on the document understanding task.

The quality of the hierarchy information has a great impact on the quality of the layout graph, which will affect the effect of GraphLayoutLM. We will explore the filed of document hierarchy layout recognition in the future.

\section{Acknowledgement}
Corresponding author: Zuchao Li and Xiantao Cai. This work was supported by the Fundamental Research Funds for the Central Universities (No.2042023kf1033) and the National Science Fund for Distinguished Young Scholars (No.62225113).

\bibliographystyle{ACM-Reference-Format}
\bibliography{sample-base}

\clearpage

\appendix

\section{Appendix}

\subsection{Pre-training Objectives}\label{appendix:layoutlmv3_pretrain}

\noindent\textbf{Masked Language Modeling (MLM)} 

To make use of the text input, we employ an unsupervised pre-training target called MLM, which is similar to BERT. We mask 30\% of the text tokens and aim to maximize the log-likelihood of the correct masked text tokens (represented as $\mathbf{y}_l$) as the pre-training objective. The text tokens are derived from corrupted sequences of image tokens ($\mathbf{X}^{I_m}$) and text tokens ($\mathbf{Y}^{L_m}$) based on their contextual representations, where $I_m$ and $L_m$ represent the masked positions. The MLM loss can be represented as follows:
\begin{equation}
L_{MLM}\left(\theta\right) = -\sum_{l=1}^{|L_m|} \log P_{\theta}\left(\mathbf{y}_\ell \mid \mathbf{X}^{I_m}, \mathbf{Y}^{L_m}\right),
\end{equation}
where $\theta$ denotes the parameters of our GraphLayoutLM model.

\noindent\textbf{Masked Image Modeling (MIM)} 

As in the patch-level MIM pre-trained ViT model~\cite{dosovitskiy2020image}, we use a blockwise masking strategy to randomly mask approximately 40\% of the image tokens. The MIM objective is also driven by cross-entropy loss, similar to MLM objective, to reconstruct the masked image tokens $x_m$ based on the context of their surrounding text and image tokens:
\begin{equation}
L_{MIM}\left(\theta\right) = -\sum_{m=1}^{|I_m|}\log P_{\theta}\left(\mathbf{x}_m \mid \mathbf{X}^{I_m}, \mathbf{Y}^{L_m}\right).
\end{equation}

\noindent\textbf{Word-Patch Alignment (WPA)}

The LayoutLMv3 paper proposes a WPA objective for learning fine-grained alignment between text words and image patches, which is useful for learning alignment between text and image. We follow this objective and use an additional two-layer Multi-Layer Perceptron (MLP) network that inputs contextual text and images and predicts binary aligned/misaligned labels. The WPA objective is implemented by a binary cross-entropy loss:
\begin{equation}
L_{WPA}\left(\theta\right) = -\sum_{\ell=1}^{L-L_m} \log P_{\theta}\left(\mathbf{z}_\ell \mid \mathbf{X}^{I_m}, \mathbf{Y}^{L_m}\right),
\end{equation}
where $L-L_m$ is the number of unmasked text tokens, $\mathbf{z}_\ell$ is the binary label of language token in the $\ell$ position.

\subsection{Finetuning Hyper-parameters}\label{appendix:parameters}

We show the finetuning hyper-parameters in Table \ref{tab:hyper_parameter}.

\begin{table}[h]
    \centering
    \small
    \caption{Finetuning hyper-parameters. L, M, B, and G refer to learning rate, max steps, batch size, and gradient accumulation steps. }
    \label{tab:hyper_parameter}
    \begin{tabular}{c|c|c|cccc}
        \toprule
        \bf Dataset & \bf \makecell[c]{Model \\ size} & \bf Language &\bf L & \bf M & \bf B & \bf G \\
        \midrule
        \multirow{2}{*}{FUNSD}
        & BASE & \multirow{2}{*}{English} & 1e-5 & 2000 & 4 & 1\\
        \multirow{2}{*}{}
        & LARGE & & 1e-5 & 2000 & 4 & 1\\
        \toprule
        \multirow{2}{*}{CORD}
        & BASE & \multirow{2}{*}{English} & 5e-5 & 4000 & 4 & 1\\
        \multirow{2}{*}{}
        & LARGE & & 1e-5 & 8000 & 4 & 1\\
        \toprule
        XFUND & BASE & CHINESE & 7e-5 & 2000 & 4 & 4 \\
        \bottomrule
    \end{tabular}
\end{table}

\subsection{Graph Reordering Strategy}\label{appendix:reordering}

\begin{figure}[h]
    \centering
    \includegraphics[width=0.9\linewidth]{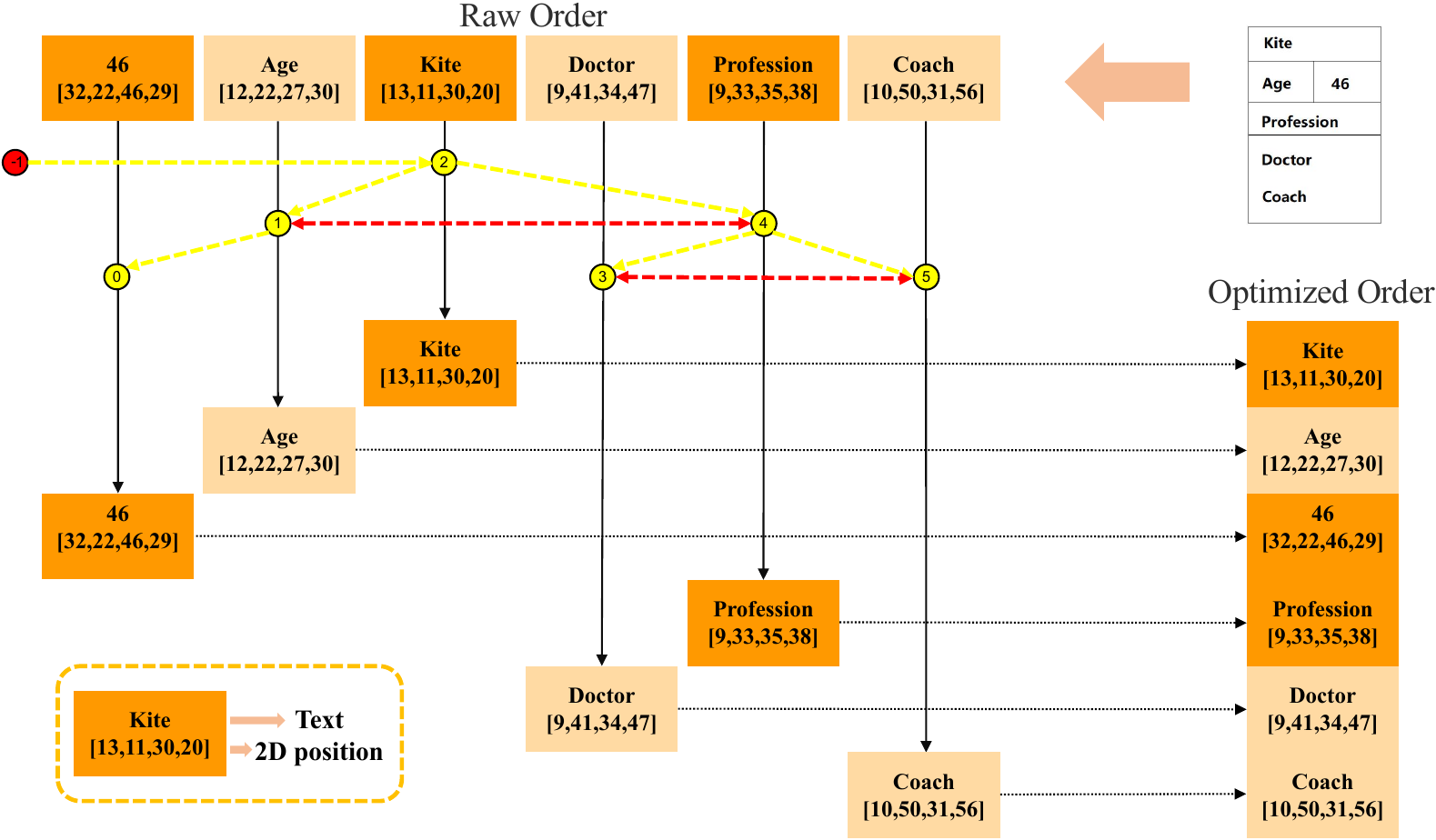}
    \centering
    \caption{An example of Graph Reordering}
    \label{fig:graph_ordering}
\end{figure}

As Figure \ref{fig:graph_ordering} shows, the raw order of document is an incomprehensible reading order. In graph reordering strategy, document structure is represented in the form of a hierarchical structure graph (the graph of yellow round nodes in Figure \ref{fig:graph_ordering}). Through the depth-first traversal method, we can sort the nodes between different levels. As the figure shown, we firstly traverse node 1 and its child nodes, and then traverse node 4.  For nodes of the same level, we use position relation to sort. For example, node 3 is at the top of node 5, so node 3 is sorted before node 5. With these two reordering rules, we achieve the reordering optimization on the raw order based on graph structure.

\subsection{Convergence Curves of the Ablation Study Fine-tuning}\label{appendix:convergence}

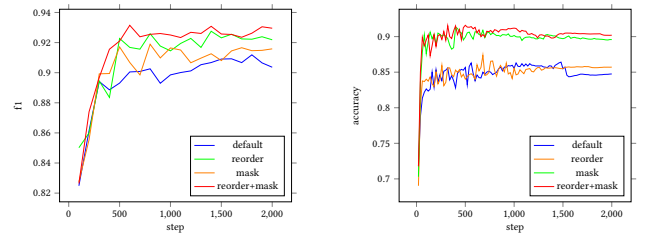
\begin{figure}  
\centering  

\begin{tikzpicture} [scale=0.45]

\begin{axis}[
    xlabel=step, 
    ylabel=f1, 
    tick align=outside, 
    legend pos=south east
    ]

\addplot[blue] table {src/funsd/funsd_f1_default.dat};
\addlegendentry{default}

\addplot[green] table {src/funsd/funsd_f1_reorder.dat};
\addlegendentry{reorder}

\addplot[orange] table {src/funsd/funsd_f1_mask.dat};
\addlegendentry{mask}

\addplot[red] table {src/funsd/funsd_f1_reorder_mask.dat};
\addlegendentry{reorder+mask}

\end{axis}
\end{tikzpicture}
\qquad
\begin{tikzpicture} [scale=0.45]

\begin{axis}[
    xlabel=step, 
    ylabel=accuracy, 
    tick align=outside, 
    legend pos=south east
    ]

\addplot[blue] table {src/xfund/xfund_accuracy_default.dat};
\addlegendentry{default}

\addplot[orange] table {src/xfund/xfund_accuracy_reorder.dat};
\addlegendentry{reorder}

\addplot[green] table {src/xfund/xfund_accuracy_mask.dat};
\addlegendentry{mask}

\addplot[red] table {src/xfund/xfund_accuracy_reorder_mask.dat};
\addlegendentry{reorder+mask}

\end{axis}
\end{tikzpicture}
\caption{Convergence Process.}
\label{fig:convergence}

\end{figure}

To increase the credibility of the experimental results, we also show the accuracy and F1 convergence curves of the ablation study fine-tuning. The line chart in the left of Figure \ref{fig:convergence} shows the convergence process of F1 on the FUNSD test set. We can see that the result with both graph reordering and graph masking methods is the best overall. Both of the results that incorporate the graph reorder and graph mask strategies respectively are overall better than the baseline. The line chart in the right of Figure \ref{fig:convergence} shows the convergence process of accuracy on XFUND. It can be seen together with Table 4 that only using the graph reordering strategy has little impact on the accuracy metric, while the graph masking strategy significantly improves accuracy.

\subsection{Results on other Datasets}\label{appendix:other_data}

We show the experiment results on SROIE and DocVQA in Table \ref{tab:result_on_other_datasets}. Because we can not successfully reproduce layoutlmv3's results on these datasets, we only show them in the appendix. The experiment results show that our proposed method performs better than LayoutLMv3.

\begin{table}[h]
    \centering
    \small
    \caption{Results on SROIE and DocVQA. }
    \label{tab:result_on_other_datasets}
    \begin{tabular}{l|c|c}
        \toprule
        \bf Methods & \bf \makecell[c]{DocVQA(ANLS)} & \bf SROIE(F1) \\
        \midrule
        LayoutLMv3-base & 72.59 & 99.25\\
        \toprule
        GraphLayoutLM-base(ours) & 73.51 & 99.30\\
        \bottomrule
    \end{tabular}
\end{table}

\subsection{The Graph Example on FUNSD}\label{appendix:graph_example}

\begin{figure}[h]
    \centering
    \includegraphics[width=0.9\linewidth]{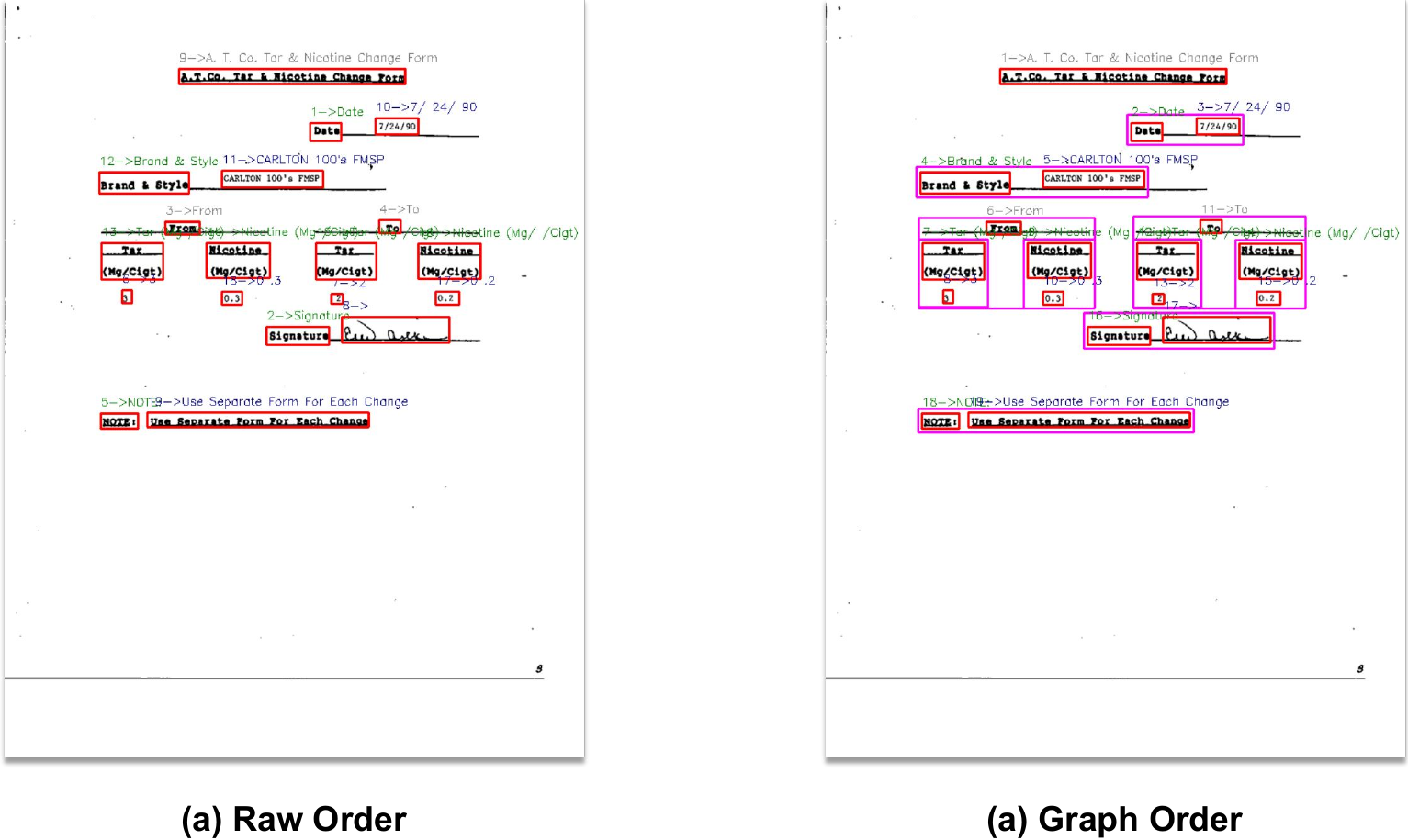}
    \centering
    \caption{An Graph Example on FUNSD}
    \label{fig:graph_example_funsd}
\end{figure}

We show the importance of graph structure with an example from FUNSD. As shown in Figure \ref{fig:graph_example_funsd}, the red box in the figure is the box of the node, and the number before the text above the box is the order. The purple box in document (b) is the domain scope of its first node. All other nodes in the purple box are children of this node. For example, "7/24/90" is a child of "Date". The position relationships are established at the same level. For example, a "left-right" relationship is established between "From" and "To" nodes.

As we can see, this graph structure represents the layout relationships in the document better. It's easy to see connections between nodes, such as "7/24/90" being the answer to "Date". Similarly, we can easily find that the order after Graph Reordering is more logical.

\subsection{Graph Models Comparison}\label{appendix:graph_model_comparison}

We compare our method with other graph structuring methods mentioned in related work. We also compare the results of our method with those of comparable models(ERNIE-mmLayout, Rope, FormNet).

Liu et al.~\cite{liu2019graph} use the GCN method and fully connect nodes in the graph, which can lead to significant time spent processing graph information. In contrast, GraphLayoutLM draws on the idea of GAT, which combines the graph structure with the attention mechanism based on GCN. Additionally, when modeling the document structure graph, we establish edge relationships based on level and position rather than full connection to strengthen node knowledge learning with edge relationships.

ERNIE-mmLayout~\cite{wang2022ernie} includes coarse-grained nodes (paragraph nodes) and fine-grained nodes (text nodes). We only use text nodes, which eliminates the need to deal with additional redundant data.Besides, compared with ERNIE-mmLayout, although the hierarchical connection manners between coarse-grained nodes and fine-grained nodes are similar, both of them adopt the tree mode. However, ERNIE-mmLayout establishes full connections between coarse-grained nodes and fine-grained nodes respectively, while our method only establishes connections between sibling nodes. As mentioned in our paper, we believe that this method containing prior knowledge can make the model pay more attention to the surrounding node relations of the same level nodes.More importantly, the graph of ERNIE-mmLayout has only two levels, and ours can be multi-level and more free (the depth of leaf nodes can be different).Finally, ERNIE-mmLayout uses a more complex network model, and we only use a multi-head self-attention layer, which makes the model lighter and faster.

ROPE~\cite{lee2021rope} sorting uses data to generate an order based on the GCN network and the method we proposed adopts the reordering rule with prior knowledge. ROPE's graph is based on the $\beta$-skeleton graph and focuses only on neighbor nodes, while our graph focuses not only on neighbor nodes but also on sibling nodes of the same level (even if they are far away).

FormNet~\cite{lee2022formnet} focuses on the surrounding information of nodes, and uses the Rich Attention part to analyze the attention of nodes and their surrounding nodes.  The graph modeling we proposed not only focuses on the peripheral information of nodes (the location information of sibling nodes of leaf nodes), but also uses hierarchical relations to construct the location relations of distant nodes (such as the relationship between paragraphs).  We believe that there may also be a strong connection between distant nodes at the same level (for example, the first and last paragraphs of an essay may have a strong coherence).

The experiments results are shown in Table \ref{tab:result_on_graph_models}, we can see that GraphLayoutLM have the best performance.

\begin{table}[h]
    \centering
    \small
    \caption{F1 Results on FUNSD and CORD. }
    \label{tab:result_on_graph_models}
    \begin{tabular}{l|c|c}
        \toprule
        \bf Methods & \bf FUNSD & \bf CORD \\
        \midrule
        ROPE &57.22 & -\\
        ERNIE-mmLayout-base & 86.02 & 97.23 \\
        GraphLayoutLM-base(ours) & \bf  93.15 & \bf  97.28\\
        \toprule
        FormNet & 84.69 & 97.28 \\
        ERNIE-mmLayout-base & 86.49 & 97.38 \\
        GraphLayoutLM-large(ours) & \bf 94.39 & \bf  94.39\\
        \bottomrule
    \end{tabular}
\end{table}

\end{sloppypar}
\end{document}